\documentclass{article}
\usepackage{spconf,amsmath,graphicx,hyperref,booktabs,caption,xcolor,multirow}

\title{How Sampling Strategy Affects Imbalance Mitigation in LiDAR Segmentation: A Study of Structured vs.\ Random Point-Based Architectures}

\name{Antonis Savva$^{\star}$ \qquad Christos Kyrkou$^{\star}$ \qquad Theocharis Theocharides$^{\star\dagger}$\thanks{This work has received funding under Grant Agreement No. 101168067, GuardAI - Enhancing Robustness and Security of Edge AI Systems for Safety-Critical Applications, with support from the European Cybersecurity Competence Centre. The views and opinions expressed are, however, those of the author(s) only and do not necessarily reflect those of the European Union or the European Cybersecurity Competence Centre. Neither the European Union nor the European Cybersecurity Competence Centre can be held responsible for them. Computational resources were provided by the High Performance Computing facility of the University of Cyprus (UCY HPC). \\
This version of the manuscript has been accepted for publication in the IEEE International Conference on Image Processing (ICIP) 2026 after peer review (Author Accepted Manuscript). It is not the final published version (Version of Record) and does not reflect any post-acceptance improvements. The Version of Record is available online at https://doi.org/10.1109/ICIP61757.2026.11630063.}}

\address{$^{\star}$ KIOS Research and Innovation Center of Excellence, and \\
        $^{\dagger}$ Department of Electrical and Computer Engineering, \\
                    University of Cyprus, 1 Panepistimiou Avenue, 2109 Aglantzia, Nicosia}

\begin{document}

\maketitle

\begin{abstract}
Class imbalance in LiDAR point clouds poses challenges for semantic segmentation in autonomous navigation and urban mapping. While 2D vision has numerous mitigation techniques, their effectiveness in 3D remains unclear. We benchmark six reweighting schemes and five imbalance-aware losses across three datasets (DALES, S3DIS, STPLS3D) using two architectures (KPConv, RandLA-Net). Inverse-frequency weighting degrades performance by up to 12\% compared to uniform weighting, with catastrophic failures in minority classes. Uniform weighting performs within 2\% of complex losses for structured sampling (KPConv) but benefits less for random sampling (RandLA-Net, up to 4.6\% gap). Loss landscape analysis reveals a complex interplay: for structured sampling, imbalance ratio determines landscape geometry on real LiDAR data but decouples from it on synthetic data; for random sampling, landscapes show high sensitivity to dataset geometry regardless of imbalance ratio. For the two evaluated point-based architectures, these results suggest that the interaction between sampling strategy (structured vs.\ random), imbalance severity, and data acquisition characteristics shapes which mitigation approaches are effective.
\end{abstract}

\begin{keywords}
LiDAR point clouds, class imbalance, loss reweighting, landscape analysis
\end{keywords}

\section{Introduction}\label{sec:intro}
LiDAR semantic segmentation is essential for autonomous navigation~\cite{Arnold2019}, urban mapping~\cite{Wang2018}, and environmental monitoring~\cite{Gaydon2024}. Unlike 2D images with regular pixel grids, LiDAR produces irregular, sparse point clouds requiring specialized network architectures. 

Class imbalance in LiDAR differs from 2D vision: (i) frequency imbalance, where majority classes (ground, buildings) outnumber minority classes (signs, pedestrians) by orders of magnitude; (ii) geometric imbalance, where points near the sensor have higher density; (iii) structural imbalance, where small objects with simple geometry are harder to distinguish.

Mitigation techniques include inverse-frequency reweighting and class-balanced sampling~\cite{Cui2019}, focal loss~\cite{Tsung-Yi2017}, label-distribution-aware margin~\cite{Cao2019}, logit adjustment~\cite{Menon2021}, Seesaw loss~\cite{Wang2021}, and balanced softmax~\cite{Ren2020}. These were developed for 2D classification (CIFAR-LT, ImageNet-LT, iNaturalist) and whether they transfer to 3D point clouds remains unclear.

Prior work shows that 3D segmentation performance depends on class frequency and geometric properties of each class~\cite{Pan2023}. However, how architectural sampling strategies interact with these factors is unexplored. KPConv~\cite{Thomas2019KPConv} uses structured, potential-based sampling, and RandLA-Net~\cite{Hu2020RandLANet} uses random sampling. These different approaches may exhibit different sensitivities to imbalance and data characteristics.

This work addresses these gaps through a systematic empirical study.  Our contributions are:
\begin{itemize}
    \item We benchmark 11 mitigation strategies, initially developed for 2D vision, on three different acquisition modalities: aerial LiDAR (DALES; 641:1 imbalance), indoor RGB-D (S3DIS; 56:1 imbalance), and photogrammetry/synthetic (STPLS3D; 101:1 imbalance).
    \item Cross-architecture evaluation using structured sampling (KPConv) and random sampling (RandLA-Net) to assess how sampling strategy affects method robustness.
    \item We show that inverse-frequency weighting consistently degrades performance (up to 12\%), while uniform weighting remains competitive for structured sampling (within 2\%) but less so for random sampling (up to 4.6\% gap).
    \item Loss landscape analysis reveals complex interactions: for structured sampling (KPConv), imbalance ratio couples to landscape geometry on real data but decouples on synthetic data; for random sampling (RandLA-Net), geometric complexity dominates landscape geometry regardless of imbalance.
\end{itemize}

\section{Related Work}\label{sec:related_work}
\subsection{3D Semantic Segmentation}
Semantic segmentation for point clouds is categorized into projection-based, voxel-based, and point-based techniques. \textbf{Projection-based} methods map 3D points onto 2D grids or spherical surfaces for CNNs but cause information loss and occlusion artifacts. \textbf{Voxel-based} methods divide space into regular grids and use 3D convolutions to capture local structure, but are memory-intensive. \textbf{Point-based} methods process raw points directly with architectures such as PointNet~\cite{Charles2017PointNet}. Advances include local feature aggregation with multilayer perceptrons \cite{Hu2020RandLANet}, kernel-point convolutions \cite{Thomas2019KPConv}, and diffusion-based segmentation~\cite{Liu2024diffusion}, which models points as particles under a probabilistic process to better reconstruct the topology.

\subsection{Class Imbalance}
Strategies for addressing class imbalance include class rebalancing (resampling, class-sensitive losses, logit adjustment), information augmentation (transfer learning, data augmentation), and module improvement (representation learning, decoupled training, ensemble methods)~\cite{Zhang2023}.

In 3D segmentation, class imbalance is particularly severe. For example, vehicle-mounted LiDAR datasets are dominated by large classes such as \textit{road}, \textit{building}, and \textit{plants}, while rare categories like \textit{people} and \textit{rider} are underrepresented \cite{Pan2023}. LiDAR sensing geometry worsens the imbalance, with objects nearer to the sensor having higher point density than those farther away, which decreases segmentation accuracy for small or distant classes.

Pan~\textit{et al.}~\cite{Pan2023} showed that 3D imbalance arises from both class frequency and intrinsic geometric properties, with geometrically similar classes (e.g., \textit{fence}/\textit{plants}) overlapping in embedding space regardless of sample count.

\section{Methodology}\label{sec:methodology}
\subsection{Reweighting Strategies}
Let \( n_i \) denote the number of points in class \( i \) where \( i = 1,\ldots,C \). We evaluate six schemes: (1) inverse logarithm \( w_i^{\text{invl}} = 1/\log(n_i) \), (2) inverse power \( w_i^{\text{invp}} = 1/n_i^{\gamma} \) with \(\gamma=0.1\), (3) complementary frequency \( w_i^{\text{comf}} = 1 - n_i/\sum_j n_j \) \cite{Prakash2023}, (4) inverse frequency \( w_i^{\text{invf}} = N/n_i \) with \( N=\sum_j n_j \), (5) class-balanced weighting \( w_i^{\text{cb}} = (1-\beta)/(1-\beta^{n_i}) \) with \(\beta=0.9\), based on the effective number of samples \cite{Cui2019}, as previously applied in \cite{Pan2023}, and (6) uniform weights \( w_i^{\text{uni}}=1\). All weights except uniform are normalized: $\sum_i w_i = 1$.

\subsection{Imbalance-Oriented Loss Functions}
We additionally benchmark five widely used imbalance-aware loss formulations:  (1) Focal Loss (FL) \cite{Tsung-Yi2017}, which down-weights easy examples: \(\mathcal{L}_{\text{FL}} = -(1-p_{y})^\gamma \log(p_y)\) with focusing parameter \(\gamma\) (we found \(\gamma=1\) to perform best in preliminary experiments); (2) LDAM \cite{Cao2019}, which enforces class-dependent margins \(\Delta_y \propto 1/n_y^{1/4}\); and (3) LADJ \cite{Menon2021}, which modifies logits as \(\tilde{z}_i = z_i - \tau \log \pi_i\) with class prior \(\pi_i = n_i/N\) (we found \(\tau=0.3\) to perform best in preliminary experiments); (4)~\textit{Balanced Softmax (BS)}, which incorporates class frequencies into the softmax calculation \cite{Ren2020}; and (5)~\textit{Seesaw Loss (SS)}, which dynamically adjusts the balance between mitigation and compensation factors, reducing penalties for tail classes and increasing penalties upon misclassification \cite{Wang2021}.

\subsection{Datasets and 3D Semantic Segmentation Models}
We used three datasets: (a) DALES (aerial LiDAR, 40 tiles, 9 classes) \cite{Varney2020}; (b) S3DIS (indoor RGB-D, 13 classes) \cite{Armeni2016}; and (c) STPLS3D (synthetic/photogrammetric, 6 classes) \cite{Chen2022}. We employ the KPConv architecture \cite{Thomas2019KPConv}, a fully convolutional network operating directly on point clouds using rigid/deformable kernels, and RandLA-Net~\cite{Hu2020RandLANet}, which processes point clouds using stacked local-feature aggregation modules and progressive random sampling. Training is conducted from scratch using the official implementations, i.e., KPConv (SGD; momentum 0.98; 400 epochs for DALES, and 500 for S3DIS and STPLS3D; initial learning rate 0.01) and RandLA-Net (Adam; 100 epochs; initial learning rate 0.01 reduced by 5\% after each epoch).

\section{Results}\label{sec:results}

\subsection{Evaluation Metrics}
We use per-class Intersection over Union (IoU):
\begin{equation} 
    \text{IoU}_i = \frac{\text{cm}_{ii}}{\text{cm}_{ii} + \sum_{j \neq i} \text{cm}_{ij} + \sum_{k \neq i} \text{cm}_{ki}}, 
    \label{eq:per_class_IoU} 
\end{equation}
where $\text{cm}_{ij}$ are confusion matrix elements. Mean IoU is: $\text{mIoU} = \frac{1}{C} \sum_{i=1}^{C} \text{IoU}_i$.

\subsection{Inverse-Frequency Weighting Harms Performance}

Table~\ref{tab:main_results} shows inverse-frequency weighting (invf) underperforms uniform across datasets. On DALES, invf degrades mIoU by 12.24\% (KPConv) and 9.39\% (RandLA-Net). On STPLS3D, degradation is 10.2\% (KPConv) and 7.8\% (RandLA-Net). Only on S3DIS with denser geometry and 56:1 imbalance does invf match uniform.

\begin{table}[t]
\begin{center}
    \resizebox{\linewidth}{!}{%
        \begin{tabular}{l|cc|cc|cc}
        \hline
        \multirow{2}{*}{Method} & \multicolumn{2}{c|}{DALES} & \multicolumn{2}{c|}{S3DIS} & \multicolumn{2}{c}{STPLS3D} \\
        & KPConv & RandLA & KPConv & RandLA & KPConv & RandLA \\
        \hline
        
        uni  & 80.014 & 76.762 & 63.120 & 61.375 & 57.093 & 53.464 \\
        \hline
        invf & \textcolor{red}{67.770} & \textcolor{red}{67.372} & 63.685 & 60.373 & \textcolor{red}{46.874} & \textcolor{red}{45.628} \\
        cb   & 78.566 & 77.068 & 63.530 & 62.406 & 58.194 & 51.650 \\
        invl & 80.682 & 76.843 & 63.704 & 61.661 & \textbf{59.284} & 48.526 \\
        invp & \textbf{80.813} & 78.481 & 62.953 & 63.686 & 56.430 & 56.607 \\
        comf & 80.618 & 76.960 & 63.568 & 62.640 & 56.710 & \textbf{58.098} \\
        \hline
        FL   & 80.114 & 75.972 & 63.173 & 62.526 & 57.447 & 49.405 \\
        LDAM & 80.655 & \textbf{79.172} & 63.286 & \textbf{64.698} & 57.797 & 52.269 \\
        LADJ & 79.627 & \textcolor{red}{66.762} & 63.954 & 64.669 & 54.711 & \textcolor{red}{41.935} \\    
        BS   & \textcolor{red}{68.159} & \textcolor{red}{66.430} & \textbf{64.752} & 63.117 & \textcolor{red}{44.155} & \textcolor{red}{45.833} \\
        SS   & 80.366 & 74.260 & 63.691 & 62.491 & 54.859 & 48.988 \\
        \hline
        \end{tabular}%
    }
\end{center}
\caption{mIoU (\%) per dataset and architecture. Best in \textbf{bold}, failures ($\geq$5\% below uniform) in \textcolor{red}{red}.}
\label{tab:main_results}
\end{table}

\begin{figure}[!b]
    \begin{minipage}[t]{0.9\linewidth}
      \centering
      \centerline{\includegraphics[width=8.5cm]{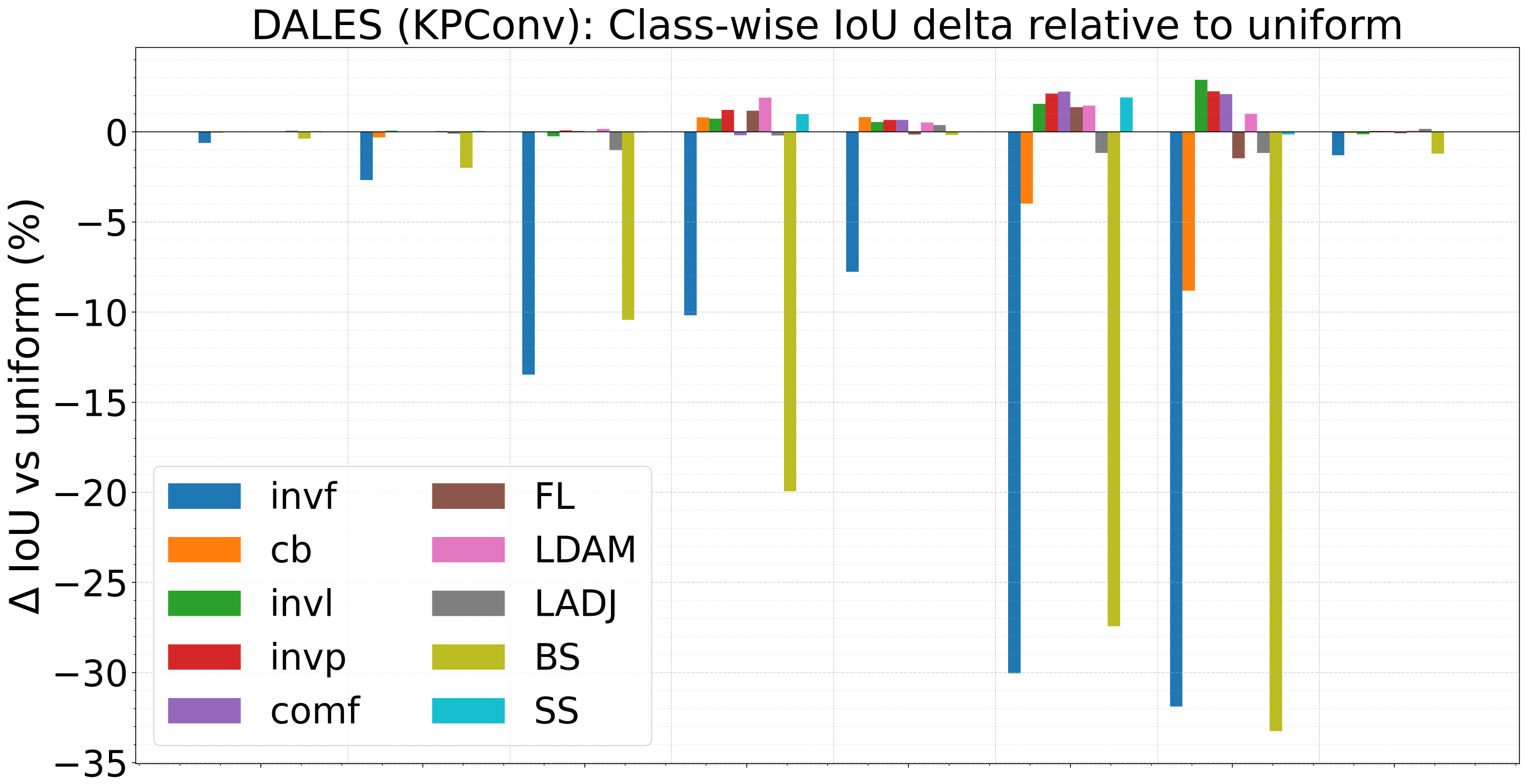}}
    \end{minipage}

    \begin{minipage}[t]{0.9\linewidth}
      \centering
      \centerline{\includegraphics[width=8.5cm]{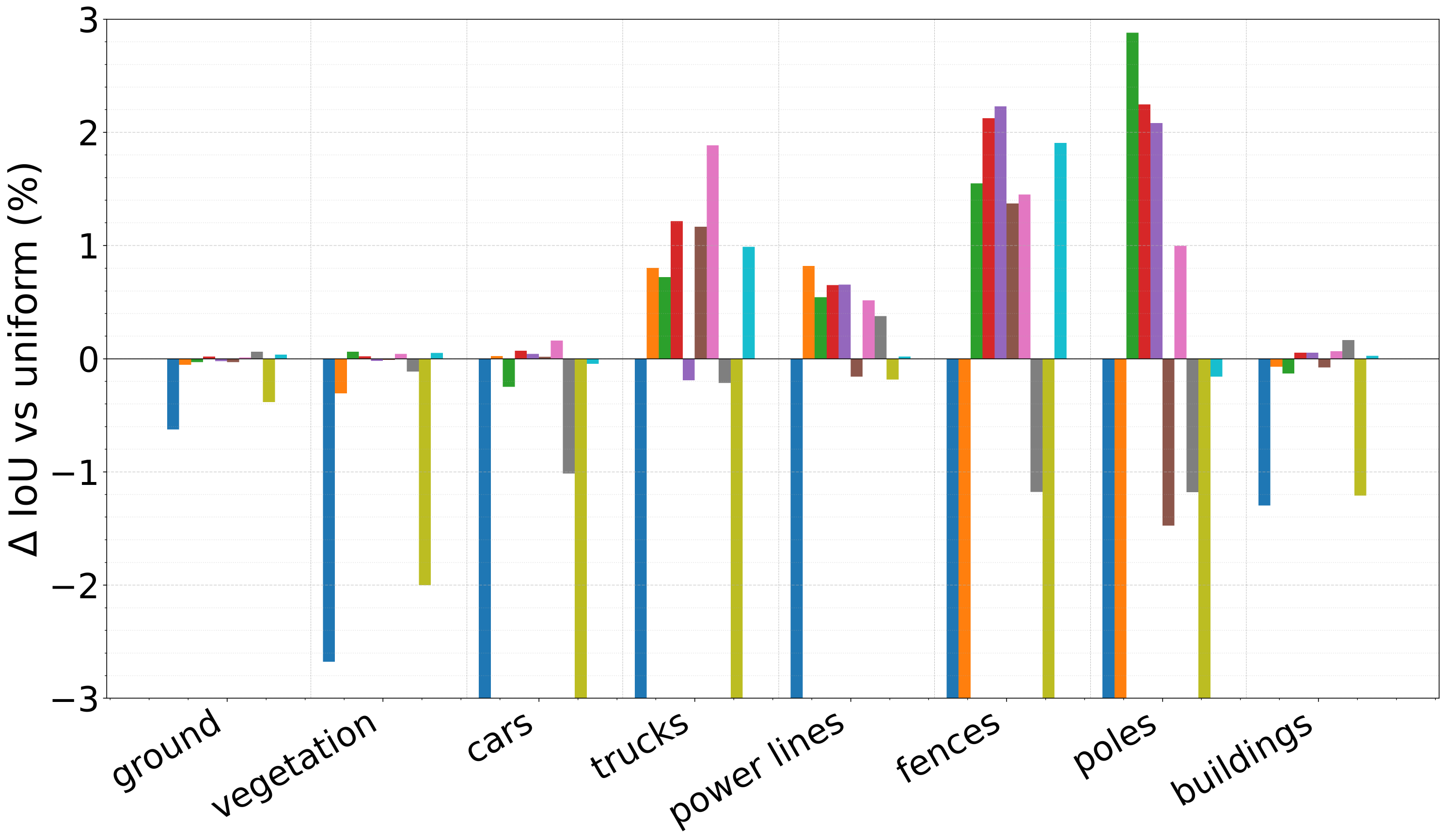}}
    \end{minipage}
    \caption{Class-wise IoU differences (`method` - `uniform`) of different methods for DALES (upper panel), and zoom-in on smaller classes (lower panel), using KPConv.}
    \label{fig:dales_kpconv_deltas}
\end{figure}

Per-class analysis (Fig.~\ref{fig:dales_kpconv_deltas}; Table \ref{tab:DALES_per_class_results_KPConv} in supplemental material) reveals the failure mechanism for DALES with KPConv (relative to uniform): \textit{cars} $-13.5\%$, \textit{trucks} $-10.2\%$, \textit{fences} $-30.0\%$ pp, \textit{poles} $-31.9\%$, with minimal gains on majority classes. Smoother reweighting (invl, invp, comf) and losses (FL, LDAM) reduce but do not eliminate this effect (Fig.~\ref{fig:dales_kpconv_deltas} lower panel). Similar patterns occur for RandLA-Net on DALES (Fig.\ref{fig:dales_randla_deltas}; Table\ref{tab:DALES_per_class_results_RandLA} in supplemental material), with the same minority classes affected most severely.

\begin{figure}[t]
    \begin{minipage}[t]{1.0\linewidth}
      \centering
      \centerline{\includegraphics[width=8.5cm]{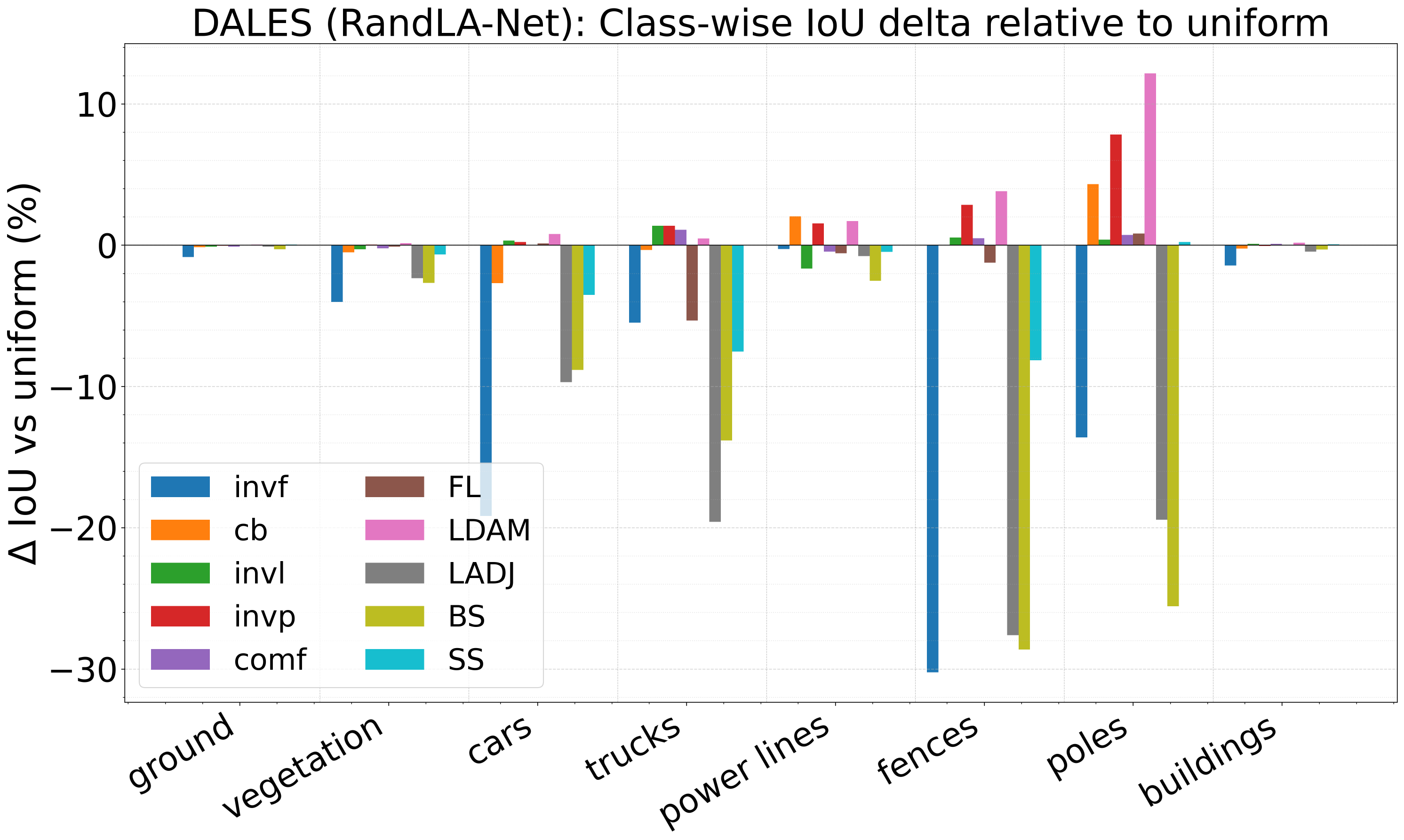}}
    \end{minipage}

    \caption{Class-wise IoU differences (`method` - `uniform`) of different methods for DALES using RandLA-Net.}
    \label{fig:dales_randla_deltas}
\end{figure}

\begin{figure}[!b]
    \begin{minipage}[t]{1.0\linewidth}
      \centering
      \centerline{\includegraphics[width=8.5cm]{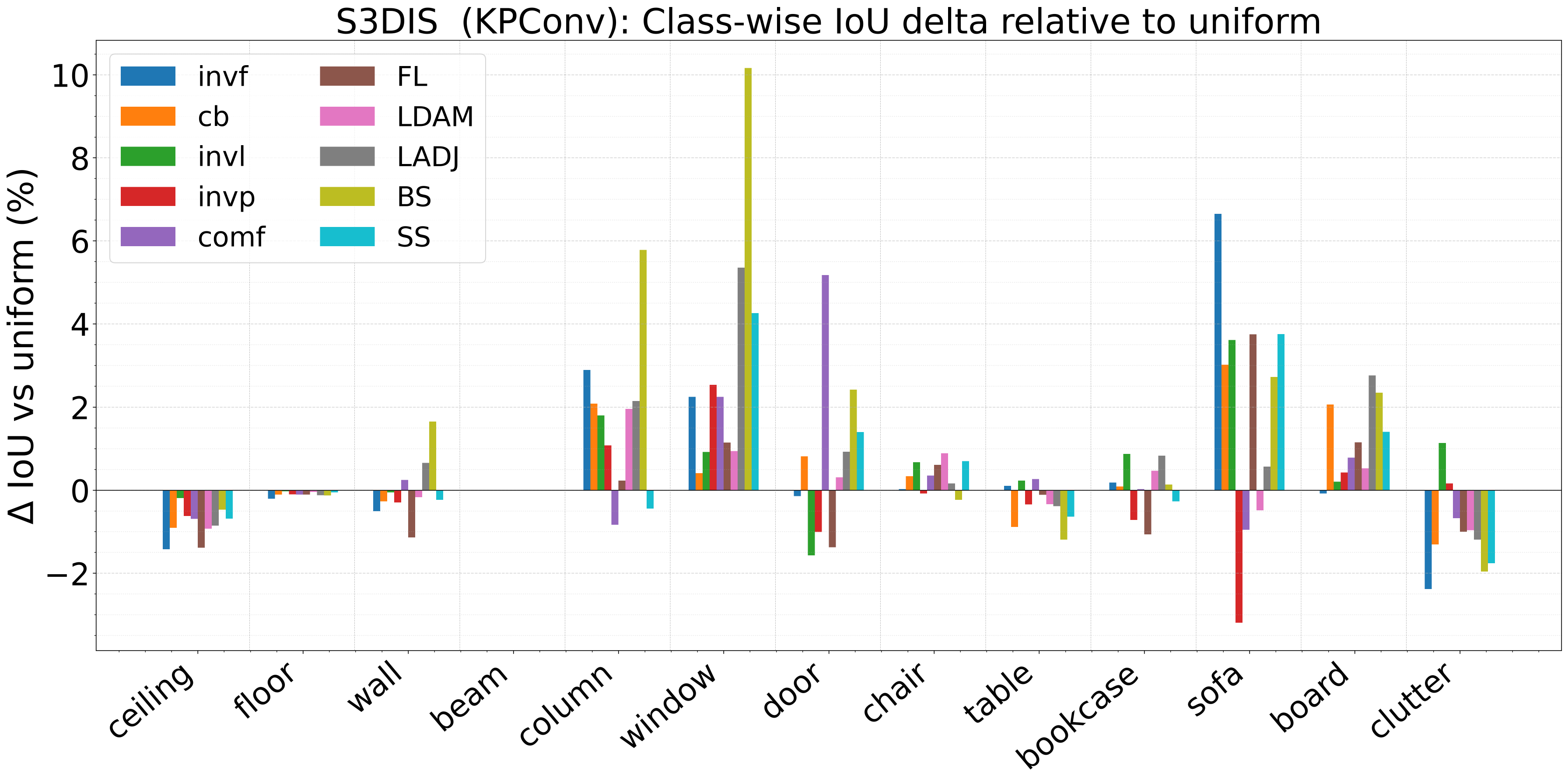}}
    \end{minipage}

    \begin{minipage}[t]{1.0\linewidth}
      \centering
      \centerline{\includegraphics[width=8.5cm]{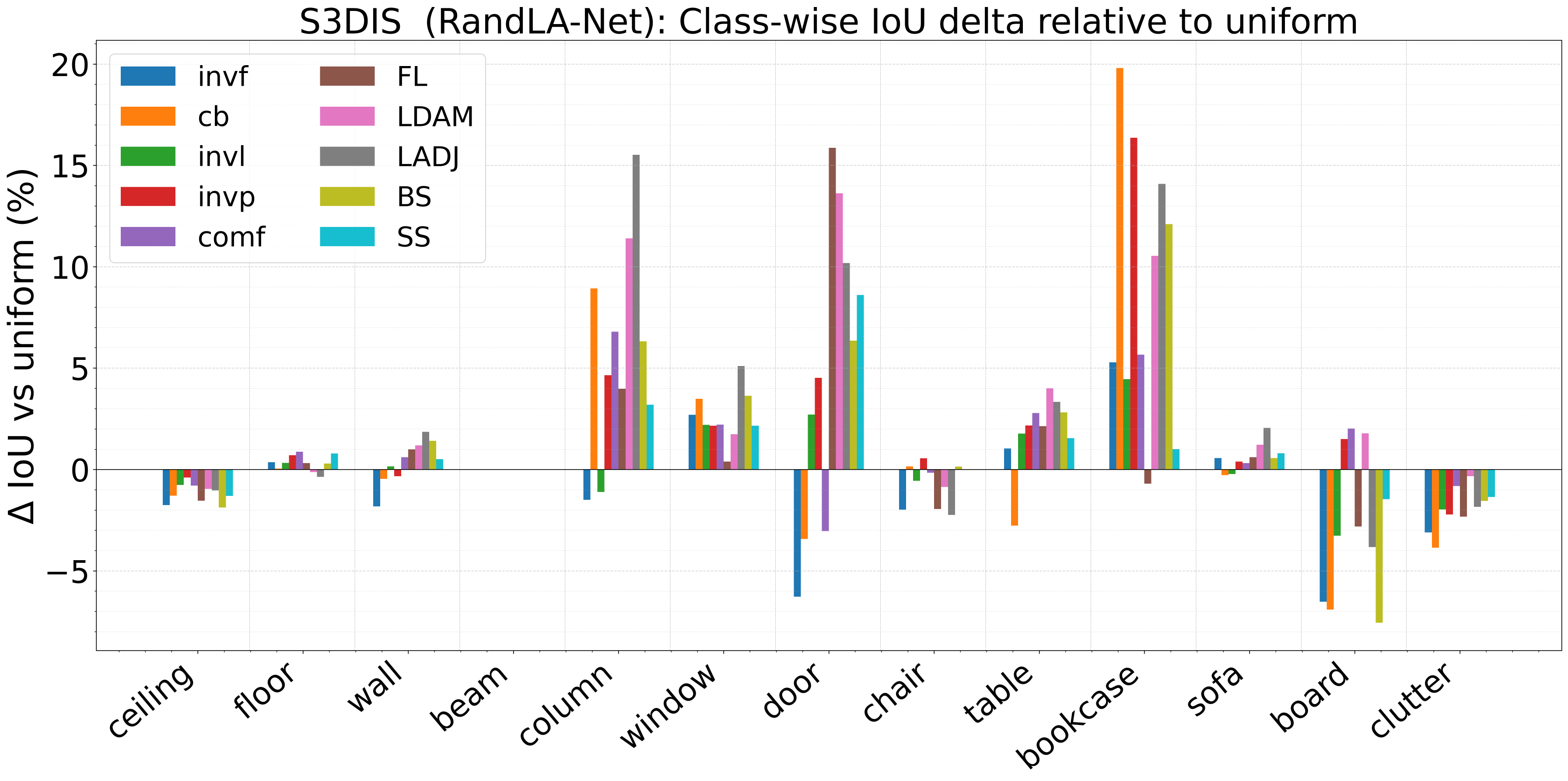}}
    \end{minipage}    
    
    \caption{Class-wise IoU differences of different methods for S3DIS using KPConv (upper panel) and RanLA-Net (lower panel).}
    \label{fig:s3dis-deltas}
\end{figure}

For S3DIS (Fig.~\ref{fig:s3dis-deltas}; Tables \ref{tab:S3DIS_per_class_results_KPConv}--\ref{tab:S3DIS_per_class_results_RandLA}), aggregate mIoU differences are compressed (62.95--64.75\% KPConv; 60.37--64.7\% RandLA-Net), yet per-class effects are substantial (e.g., BS: \textit{window} +10.2\%; cb: \textit{bookcase} +19.8\% on RandLA-Net), demonstrating that aggregate mIoU can mask opposing per-class trends.

\begin{figure}[t]
    \begin{minipage}[t]{1.0\linewidth}
      \centering
      \centerline{\includegraphics[width=8.5cm]{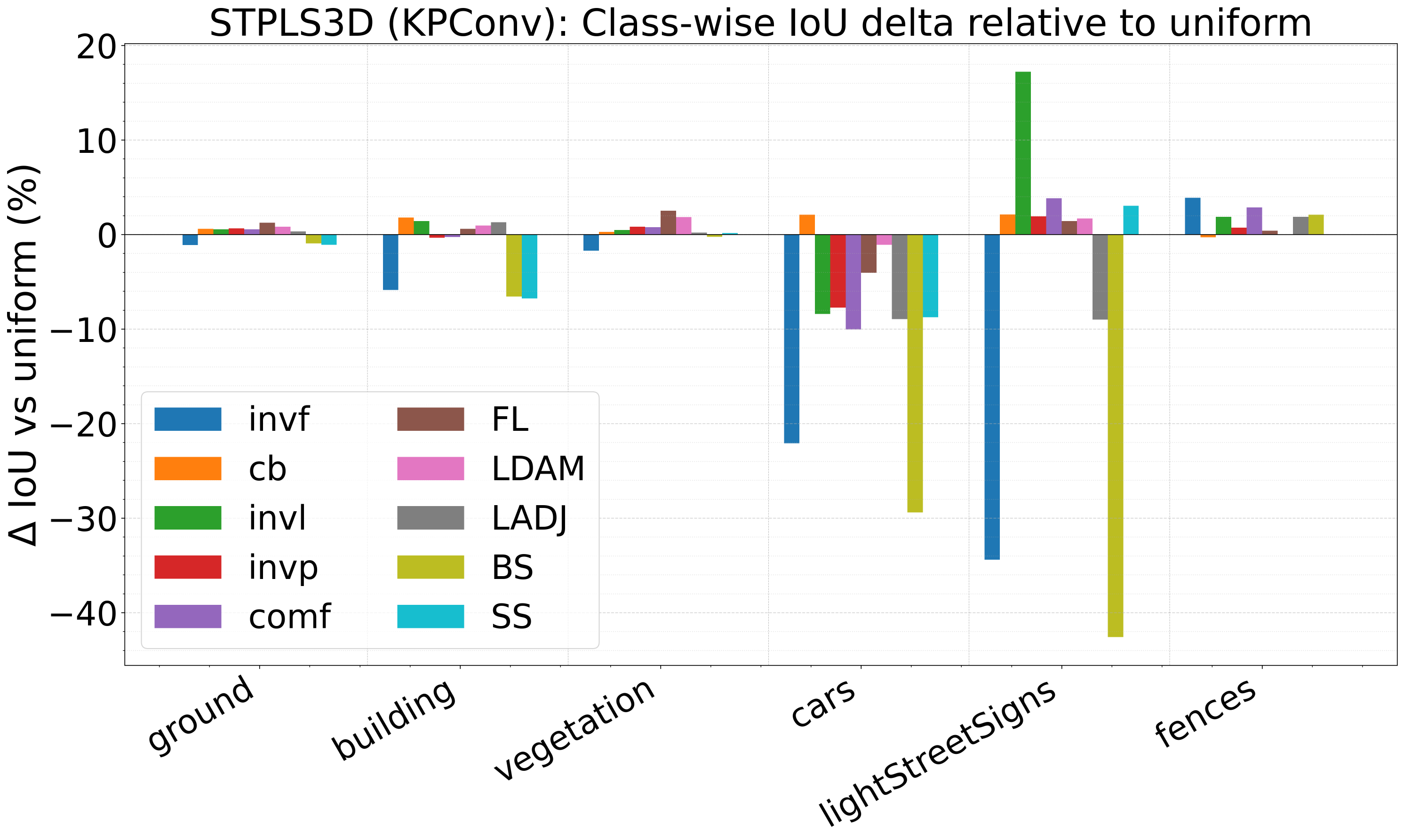}}
    \end{minipage}

    \begin{minipage}[t]{1.0\linewidth}
      \centering
      \centerline{\includegraphics[width=8.5cm]{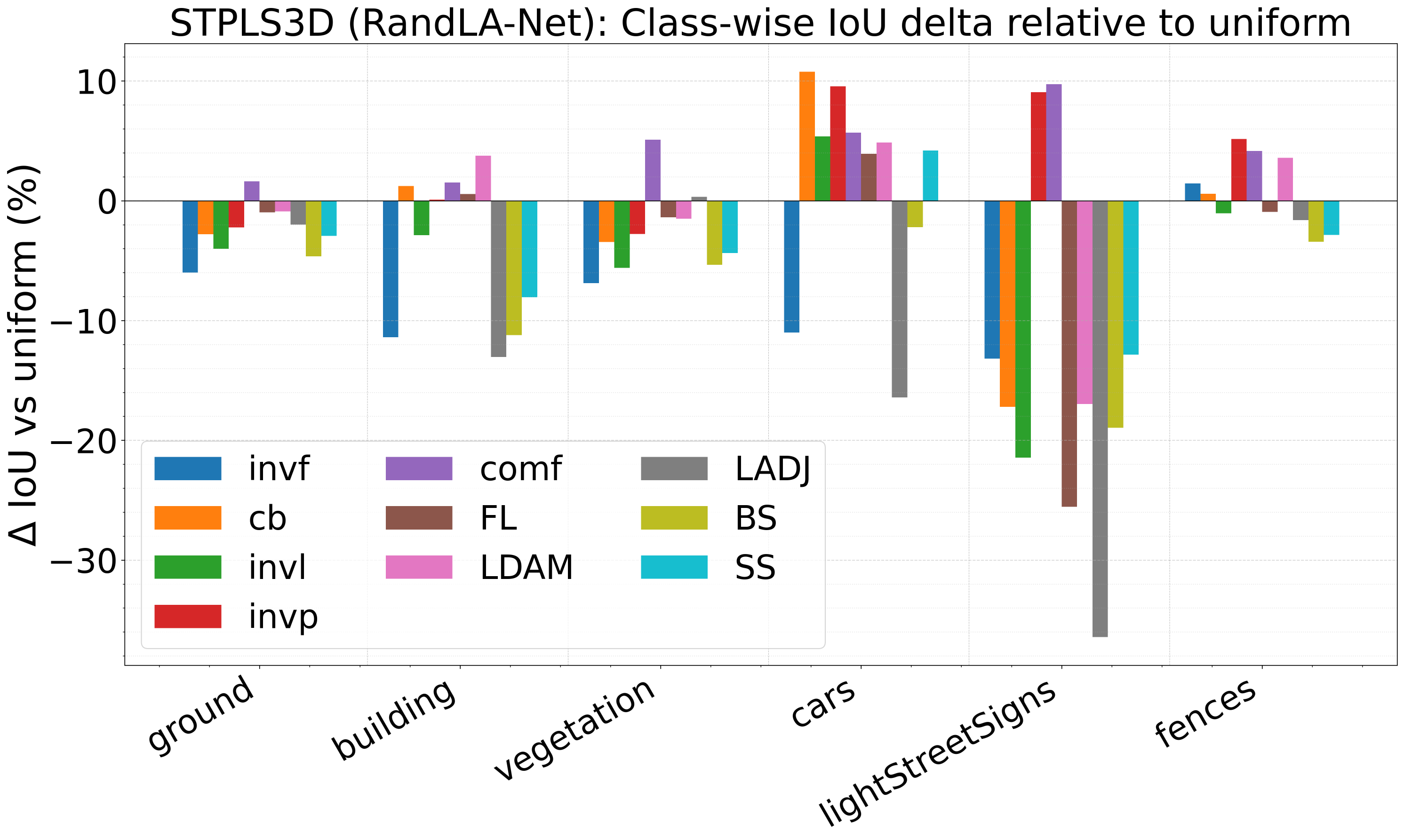}}
    \end{minipage}
    
    \caption{Class wise IoU differences of different methods for STPLS3D using KPConv (upper panel) and RanLA-Net (lower panel).}
    \label{fig:STPLS3D-deltas}
\end{figure}

On STPLS3D (Fig.~\ref{fig:STPLS3D-deltas}; Tables \ref{tab:STPLS3D_per_class_results_KPConv}--\ref{tab:STPLS3D_per_class_results_RandLA}), patterns differ between architectures: invl is best for KPConv (\textit{light/street signs} +17.2\%), comf for RandLA-Net (\textit{light/street signs} +9.7\%, \textit{fences} +4.2\%). Invf, LADJ, and BS show severe degradation, confirming that aggressive logit-based reweighting poses risks.

\subsection{Uniform Weighting is Competitive for Structured Sampling}

For KPConv, uniform weighting performs within 2.2\% of the best method across all datasets. On DALES with 641:1 imbalance, uniform achieves 80.01\% mIoU versus 80.81\% for the best (invp). For RandLA-Net, the gap widens, i.e., uniform trails the best by up to 4.6\% (STPLS3D), and specialized losses provide larger gains (LDAM: +2.4\% on DALES, +3.3\% on S3DIS).

On S3DIS, differences compress for both architectures (mIoU range: 62.95--64.75\% for KPConv; 60.37--64.7\% for RandLA-Net), suggesting dense geometry with moderate imbalance reduces the importance of loss function engineering.

\subsection{Sampling Strategy Modulates Sensitivity to Loss Design}
For KPConv, optimal loss selection gains at most 2.2\%. For RandLA-Net, gains are larger but some losses fail catastrophically. Both LADJ and BS adjust logits using class frequencies before softmax, yet behave differently: LADJ is tolerated by KPConv but fails on RandLA-Net ($-10.0$\% on DALES, $-11.5$\% on STPLS3D), while BS degrades on both architectures under high imbalance (up to $-12.9$\% on STPLS3D). This suggests that logit-adjustment formulations are inherently fragile under severe imbalance, with structured sampling partially mitigating instability for LADJ but not for BS.

\begin{figure}[t]
    \begin{minipage}[t]{1.0\linewidth}
      \centering
      \centerline{\includegraphics[width=8.5cm]{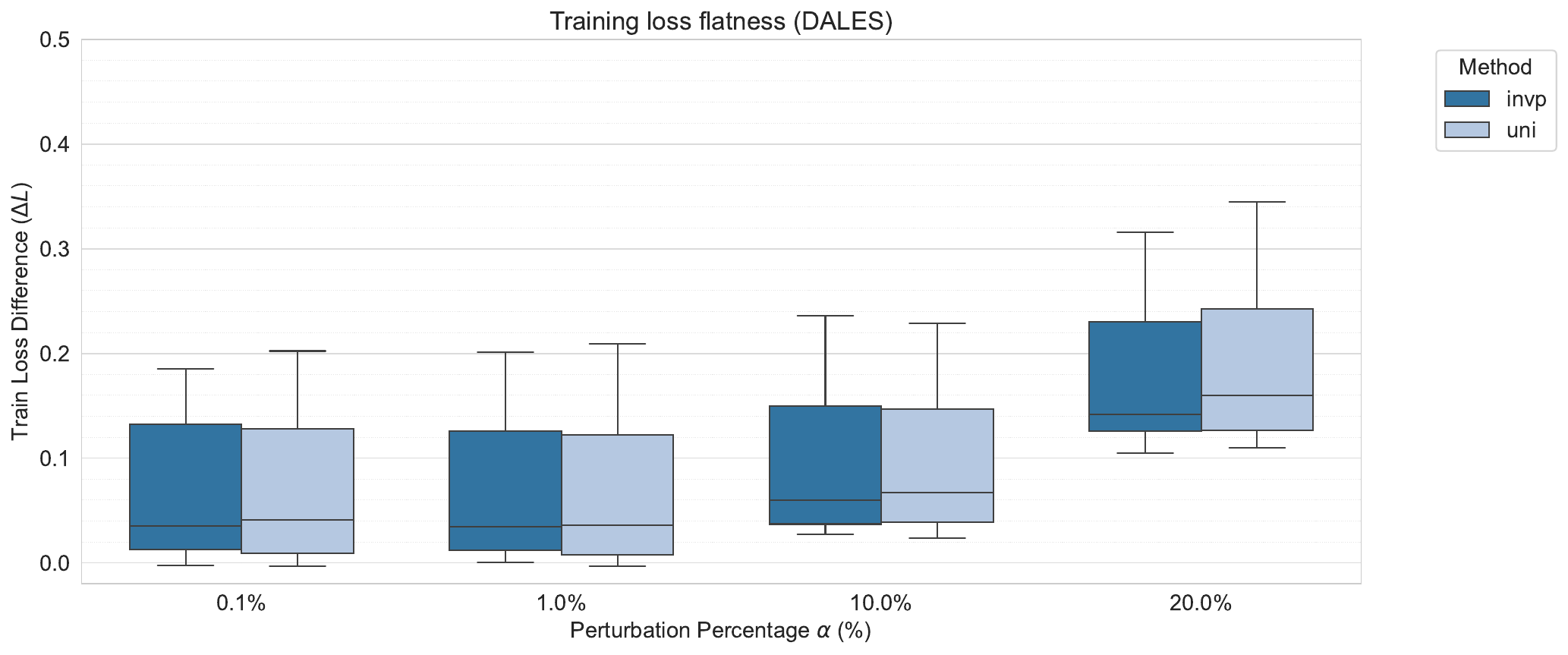}}
    \end{minipage}

    \begin{minipage}[t]{1.0\linewidth}
      \centering
      \centerline{\includegraphics[width=8.5cm]{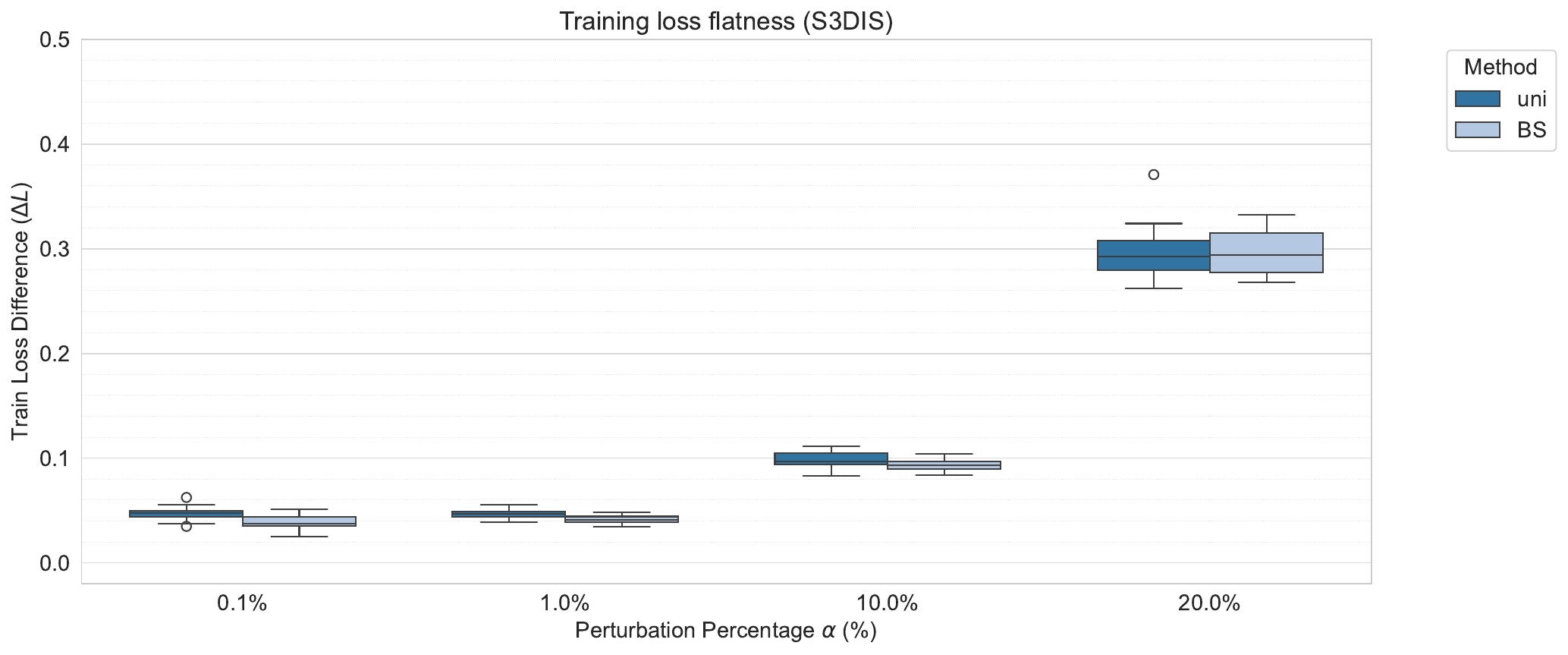}}
    \end{minipage}

    \begin{minipage}[t]{1.0\linewidth}
      \centering
      \centerline{\includegraphics[width=8.5cm]{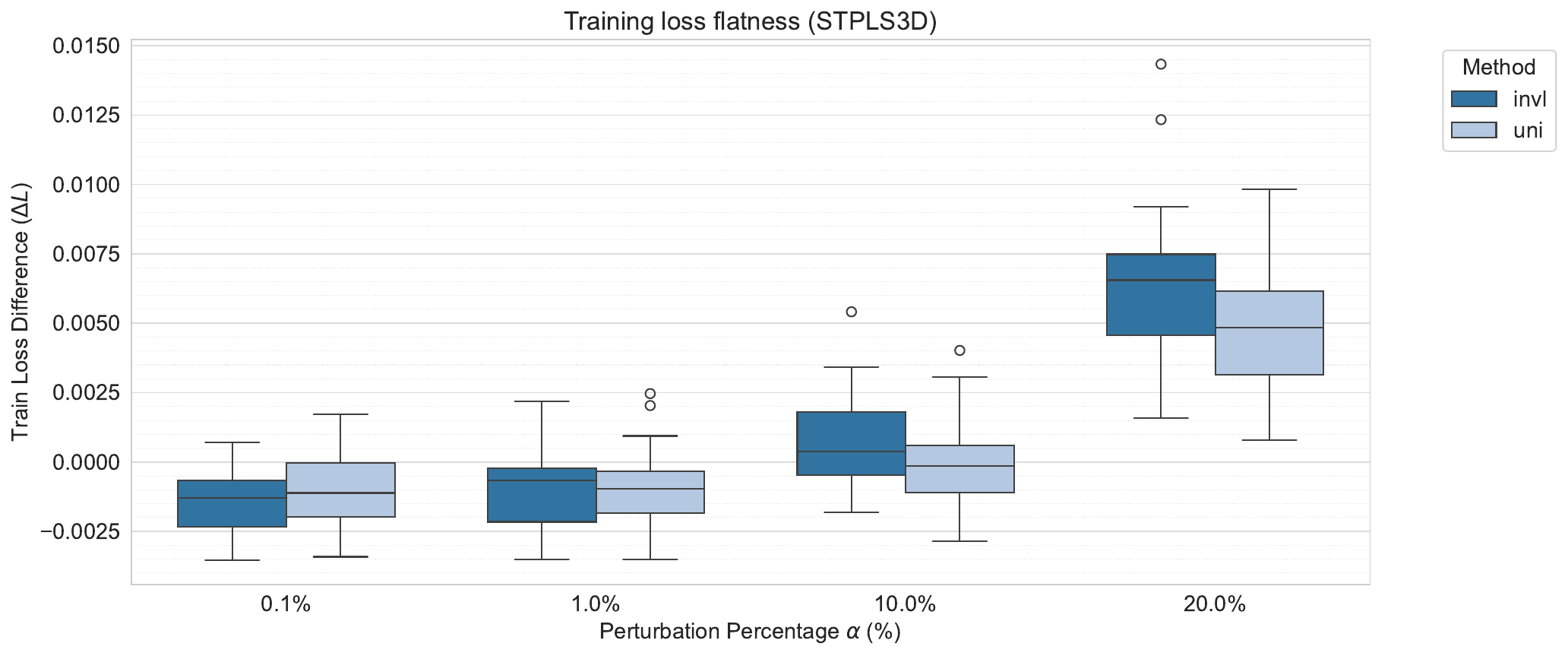}}
    \end{minipage}    
    
    \caption{Loss landscape flatness analysis for KPConv on DALES, S3DIS, and STPLS3D. The plots show the loss deviation $\delta$ as weights are perturbed with magnitude $\alpha$.}
    \label{fig:flatness_kpconv}
\end{figure}

\subsection{Loss Landscape Analysis}
We analyze loss landscape flatness at converged $\theta^*$ using filter-normalized perturbations~\cite{Li2018}:
\begin{equation}
\delta = \mathcal{L}_{\text{train}}(\theta^* + \alpha \hat{v}) - \mathcal{L}_{\text{train}}(\theta^*),
\end{equation}
where $\alpha$ is perturbation magnitude and $\hat{v}$ is a random direction. Smaller $\delta$ indicates flatter geometry.

\textbf{Structured sampling.}
For KPConv (Fig.~\ref{fig:flatness_kpconv}), comparing the two imbalance extremes (DALES and S3DIS) shows that imbalance ratio affects landscape geometry, where DALES (641:1) exhibits higher variance and sharper minima than S3DIS (56:1). However, STPLS3D (101:1) shows the flattest landscape despite being more imbalanced than S3DIS. This suggests that for structured sampling on real LiDAR data (DALES, S3DIS), imbalance ratio correlates with landscape sharpness, but synthetic/photogrammetric data (STPLS3D) decouples this relationship, i.e., clean boundaries and reduced label noise dominate over imbalance. In all cases, uniform weighting performs within 2.2\% of the best method.

\begin{figure}[t]
    \begin{minipage}[t]{1.0\linewidth}
      \centering
      \centerline{\includegraphics[width=8.5cm]{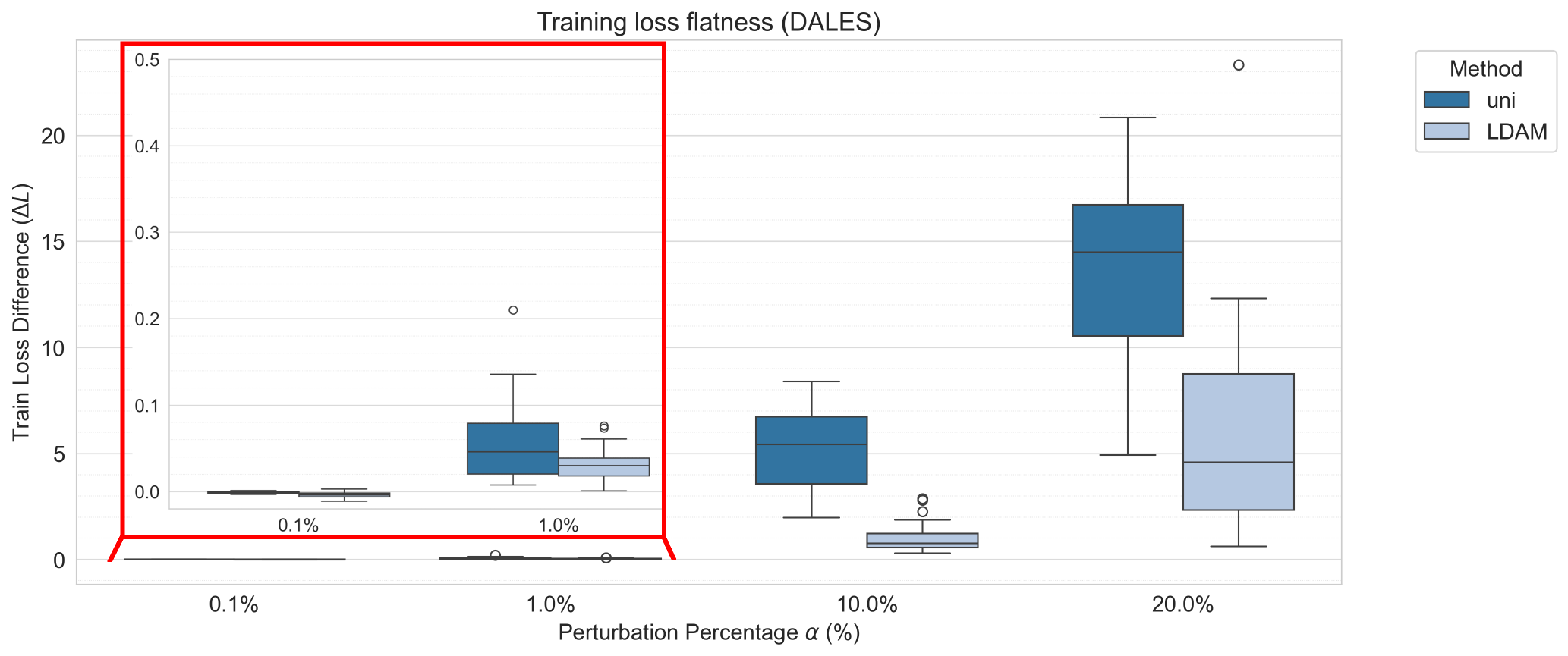}}
    \end{minipage}

    \begin{minipage}[t]{1.0\linewidth}
      \centering
      \centerline{\includegraphics[width=8.5cm]{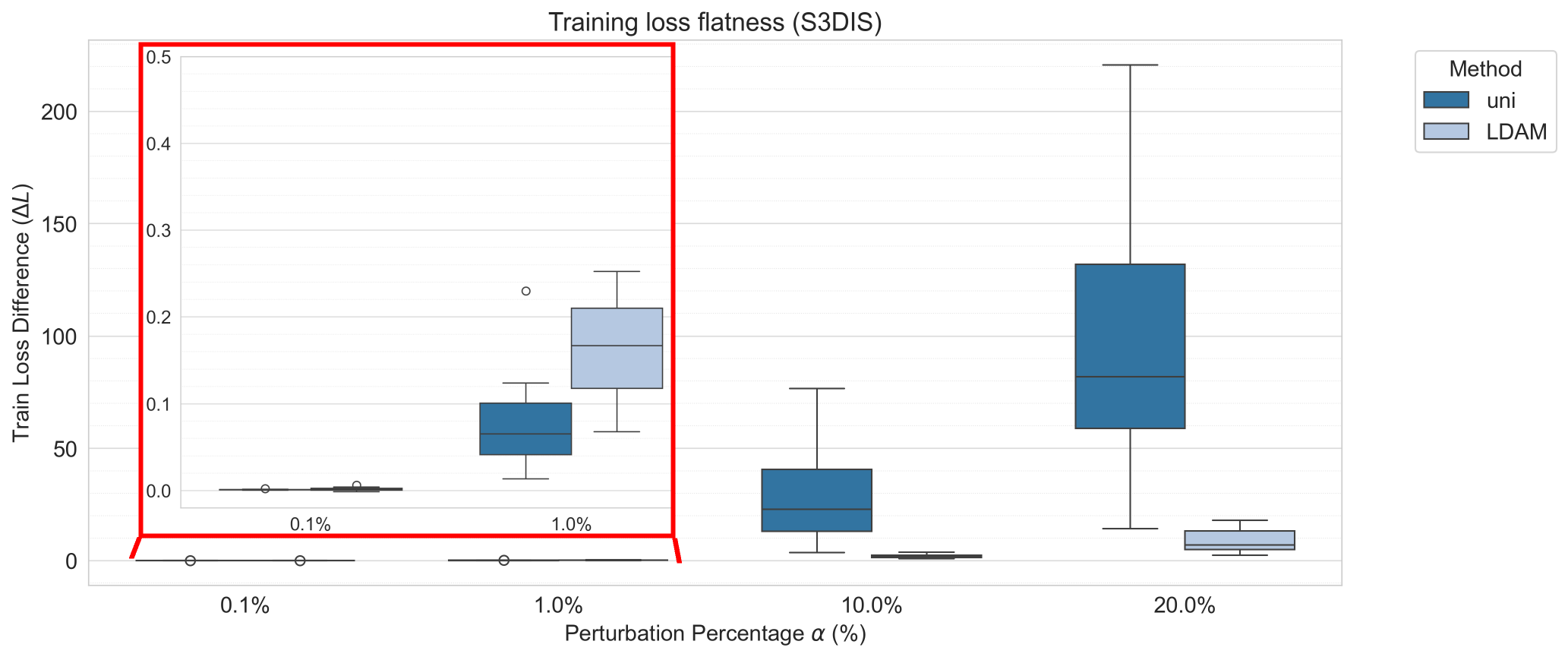}}
    \end{minipage}

    \begin{minipage}[t]{1.0\linewidth}
      \centering
      \centerline{\includegraphics[width=8.5cm]{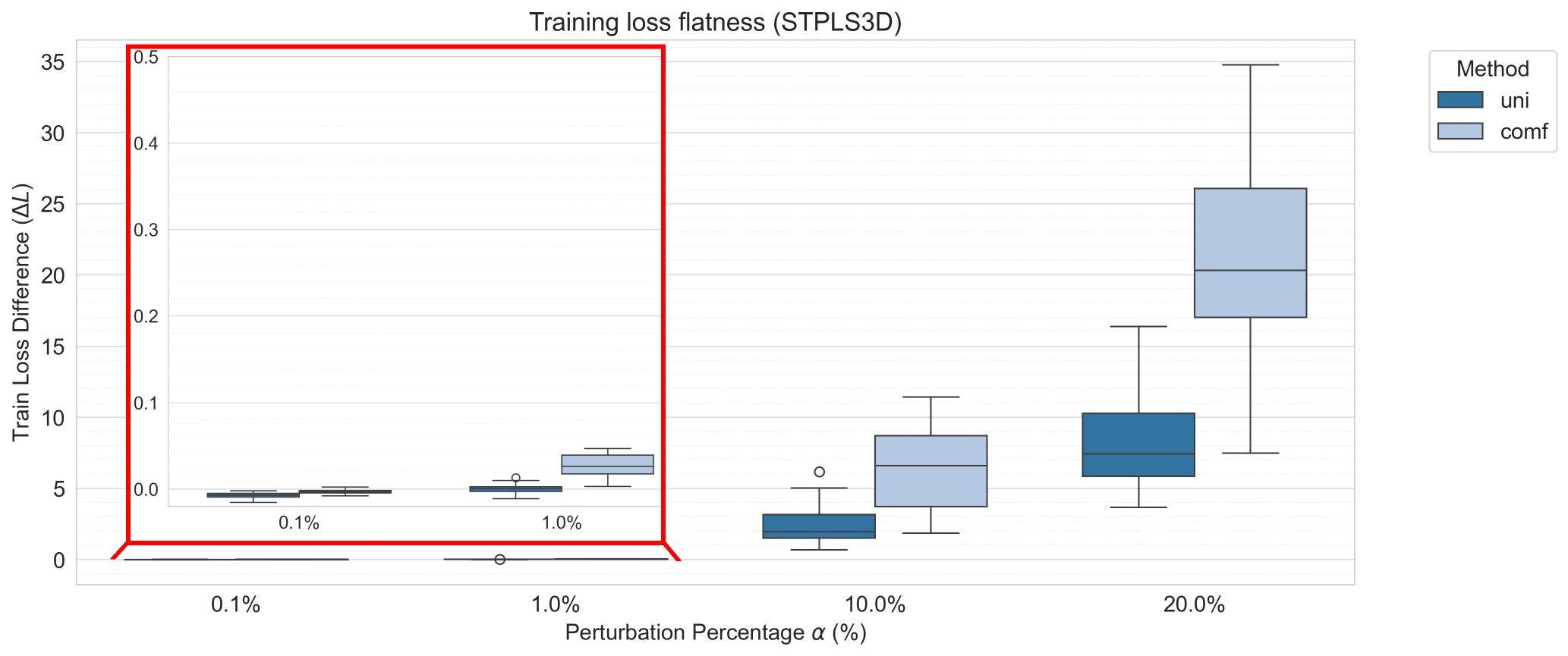}}
    \end{minipage}    
    
    \caption{Loss landscape flatness analysis for RandLA on DALES, S3DIS, and STPLS3D. The plots show the loss deviation $\delta$ as weights are perturbed with magnitude $\alpha$. Figure insets (red boxes) provide a zoomed view of low perturbation magnitudes.}
    \label{fig:flatness_randla}
\end{figure}

\textbf{Random sampling.} RandLA-Net (Fig.~\ref{fig:flatness_randla}) behaves differently. At small perturbations (0.1--1.0\%), landscapes are relatively flat across datasets. At large perturbations (10--20\%), S3DIS shows the sharpest landscape despite having the lowest imbalance (56:1), with $\delta \sim 200$ versus $\sim 20$ for DALES and $\sim 30$ for STPLS3D. This counter-intuitive pattern suggests that for random sampling, geometric complexity (dense indoor clutter with thin structures such as \textit{chairs}, \textit{tables}, \textit{columns}) creates optimization challenges that exceed those from imbalance alone. Random sampling discards geometric structure that structured sampling preserves, amplifying sensitivity to scene complexity. Here, uniform weighting trails behind by up to 4.6\%, suggesting specialized losses provide more benefit.

These patterns reveal a two-way interaction: (i)~for structured sampling on real LiDAR, imbalance ratio determines landscape geometry, but on synthetic data, quality dominates; (ii)~for random sampling, geometric complexity dominates regardless of imbalance ratio, explaining why uniform weighting suffices for KPConv but specialized losses benefit RandLA-Net.

\section{Conclusions}\label{conclusions}
We evaluated 11 class imbalance mitigation strategies across three LiDAR datasets and two point-based architectures, i.e., KPConv (structured sampling) and RandLA-Net (random sampling). For the two evaluated architectures, the interaction between sampling strategy, imbalance severity, and data acquisition characteristics shapes which mitigation approaches are effective; these observations are specific to KPConv and RandLA-Net and should be validated on additional paradigms before broader generalization.

We hypothesize that geometric structure preservation during sampling drives the observed differences: structured sampling retains local fidelity, potentially providing implicit balancing through neighborhood aggregation, while random sampling discards structure, amplifying sensitivity to scene complexity and imbalance. Attention-based architectures may exhibit intermediate sensitivity, a testable prediction we leave to future work.

Based on the empirical findings (Table~\ref{tab:main_results}, Figs.~\ref{fig:flatness_kpconv}--\ref{fig:flatness_randla}), we offer practical recommendations: (1)~avoid inverse-frequency weighting regardless of architecture or dataset (consistent degradation up to 12\%); (2)~for structured-sampling architectures (KPConv), uniform weighting is a robust default across all datasets, with smoother reweighting (invl, invp) yielding marginal gains on real LiDAR (+0.8\%); (3)~for random-sampling architectures (RandLA-Net), LDAM is the most robust specialized loss, improving over uniform on real LiDAR and RGB-D data (+2.4--3.3\%), while on photogrammetric data smoother reweighting (comf) is more effective (+4.6\%). LADJ and BS are brittle, failing catastrophically on high-imbalance data (DALES, STPLS3D) despite showing gains under moderate imbalance (S3DIS); (4)~for structured sampling on synthetic/photogrammetric data, data quality appears to dominate over loss-function choice (Figs.~\ref{fig:flatness_kpconv}--\ref{fig:flatness_randla}). Future work should investigate combined data-level and loss-level strategies, geometry-aware augmentation, and validation with voxel-based and attention-based architectures.

\bibliographystyle{IEEEbib}
\bibliography{refs}

\clearpage

\appendix

\renewcommand{\thesection}{S\arabic{section}}
\setcounter{figure}{0}
\renewcommand{\thefigure}{S\arabic{figure}}
\setcounter{table}{0}
\renewcommand{\thetable}{S\arabic{table}}

\setcounter{page}{1}

\twocolumn[
\begin{center}
    {\Large \textbf{How Sampling Strategy Affects Imbalance Mitigation in LiDAR Segmentation: A Study of Structured vs.\ Random Point-Based Architectures} \\
    \vspace{0.5em}Supplemental Material \\
    \vspace{1.0em}}
\end{center}
]

\begin{minipage}[t]{0.48\textwidth}
    \centering
    \resizebox{\linewidth}{!}{%
        \begin{tabular}{l|cccccccc}
        \toprule
        Method & \textit{ground} & \textit{vegetation} & \textit{cars} & \textit{trucks} & \textit{power lines} & \textit{fences} & \textit{poles} & \textit{buildings} \\
        \midrule
        uni     & 96.510 & 93.779 & 85.062 & 42.676 & 93.960 & 61.052 & 72.155 & 94.920 \\
        invf    & 95.883 & 91.102 & 71.594 & 32.495 & 86.191 & 31.007 & 40.264 & 93.622 \\
        cb      & 96.456 & 93.471 & 85.084 & 43.479 & 94.779 & 57.073 & 63.340 & 94.849 \\
        invl    & 96.480 & 93.841 & 84.812 & 43.398 & 94.503 & 62.600 & 75.034 & 94.788 \\
        invp    & 96.529 & 93.799 & 85.131 & 43.889 & 94.611 & 63.175 & 74.402 & 94.972 \\
        comf    & 96.488 & 93.758 & 85.105 & 42.484 & 94.615 & 63.281 & 74.237 & 94.972 \\
        \hline
        FL      & 96.480 & 93.768 & 85.078 & 43.840 & 93.801 & 62.423 & 70.678 & 94.843 \\
        LDAM    & 96.520 & 93.821 & 85.223 & 44.560 & 94.475 & 62.502 & 73.150 & 94.986 \\
        LADJ    & 96.571 & 93.665 & 84.047 & 42.462 & 94.336 & 59.874 & 70.975 & 95.084 \\       
        BS      & 96.126 & 91.777 & 74.630 & 22.739 & 93.775 & 33.615 & 38.903 & 93.710 \\
        SS      & 96.546 & 93.830 & 85.016 & 43.662 & 93.979 & 62.957 & 71.996 & 94.944 \\
        \bottomrule
        \end{tabular}%
    }
    \captionof{table}{Per-class IoU (\%) for the DALES dataset using KPConv.}
    \label{tab:DALES_per_class_results_KPConv}
    
    \vspace{3em}
    
    \resizebox{\linewidth}{!}{%
        \begin{tabular}{l|cccccccc}
        \toprule
        Method & \textit{ground} & \textit{vegetation} & \textit{cars} & \textit{trucks} & \textit{power lines} & \textit{fences} & \textit{poles} & \textit{buildings} \\
        \midrule
        uni     & 97.148 & 93.463 & 83.352 & 38.220 & 91.493 & 53.511 & 60.266 & 96.641 \\
        invf    & 96.313 & 89.435 & 64.182 & 32.725 & 91.212 & 23.269 & 46.648 & 95.193 \\
        cb      & 97.007 & 92.955 & 80.660 & 37.879 & 93.530 & 53.537 & 64.579 & 96.401 \\
        invl    & 97.034 & 93.175 & 83.681 & 39.589 & 89.842 & 54.049 & 60.650 & 96.726 \\
        invp    & 97.109 & 93.499 & 83.569 & 39.585 & 93.032 & 56.368 & 68.105 & 96.577 \\
        comf    & 97.035 & 93.232 & 83.347 & 39.306 & 91.040 & 54.002 & 60.994 & 96.728 \\
        \hline
        FL      & 97.155 & 93.356 & 83.481 & 32.884 & 90.917 & 52.261 & 61.089 & 96.633 \\
        LDAM    & 97.182 & 93.603 & 84.144 & 38.690 & 93.199 & 57.326 & 72.420 & 96.809 \\
        LADJ    & 97.038 & 91.130 & 73.658 & 18.631 & 90.722 & 25.906 & 40.825 & 96.186 \\
        BS      & 96.852 & 90.782 & 74.525 & 24.390 & 88.973 & 24.881 & 34.702 & 96.333 \\
        SS      & 97.192 & 92.805 & 79.831 & 30.681 & 91.024 & 45.357 & 60.493 & 96.694 \\
        \bottomrule
        \end{tabular}%
    }
    \captionof{table}{Per-class IoU (\%) for the DALES dataset using RandLA-Net.}
    \label{tab:DALES_per_class_results_RandLA}
\end{minipage}%
\hfill
\begin{minipage}[t]{0.48\textwidth}
    \centering
    \resizebox{\linewidth}{!}{%
        \begin{tabular}{l|cccccc}
        \toprule
        Method & \textit{ground} & \textit{building} & \textit{vegetation} & \textit{cars} & \textit{lightStreetSigns} & \textit{fences} \\
        \midrule
        uni   & 86.493 & 80.128 & 66.169 & 49.832 & 49.189 & 10.748 \\
        invf  & 85.371 & 74.248 & 64.457 & 27.737 & 14.798 & 14.630 \\
        cb    & 87.103 & 81.918 & 66.452 & 51.929 & 51.307 & 10.458 \\
        invl  & 87.052 & 81.551 & 66.655 & 41.436 & 66.402 & 12.607 \\
        invp  & 87.137 & 79.778 & 67.001 & 42.094 & 51.101 & 11.468 \\
        comf  & 87.047 & 79.856 & 66.938 & 39.793 & 53.022 & 13.603 \\
        \hline
        FL    & 87.737 & 80.737 & 68.685 & 45.781 & 50.606 & 11.138 \\
        LDAM  & 87.317 & 81.082 & 68.016 & 48.747 & 50.894 & 10.724 \\
        LADJ  & 86.809 & 81.431 & 66.363 & 40.872 & 40.175 & 12.614 \\  
        BS    & 85.546 & 73.559 & 65.934 & 20.434 & 6.612  & 12.844 \\
        SS    & 85.399 & 73.363 & 66.313 & 41.072 & 52.228 & 10.779 \\
        \bottomrule
        \end{tabular}%
    }
    \captionof{table}{Per-class IoU (\%) for the STPLS3D dataset using KPConv.}
    \label{tab:STPLS3D_per_class_results_KPConv}
    
    \vspace{1em}
    
    \resizebox{\linewidth}{!}{%
        \begin{tabular}{l|cccccc}
        \toprule
        Method & \textit{ground} & \textit{building} & \textit{vegetation} & \textit{cars} & \textit{lightStreetSigns} & \textit{fences} \\
        \midrule
        uni   & 85.007 & 76.260 & 66.441 & 40.978 & 44.774 & 7.322 \\
        invf  & 79.011 & 64.866 & 59.558 & 29.967 & 31.597 & 8.769 \\
        cb    & 82.205 & 77.489 & 62.999 & 51.747 & 27.566 & 7.895 \\
        invl  & 80.987 & 73.378 & 60.836 & 46.349 & 23.336 & 6.267 \\
        invp  & 82.773 & 76.344 & 63.670 & 50.526 & 53.845 & 12.484 \\
        comf  & 86.634 & 77.786 & 71.537 & 46.661 & 54.496 & 11.472 \\
        \hline
        FL    & 84.037 & 76.817 & 65.055 & 44.904 & 19.221 & 6.396 \\
        LDAM  & 84.116 & 80.019 & 64.938 & 45.839 & 27.797 & 10.906 \\
        LADJ  & 83.011 & 63.219 & 66.775 & 24.551 & 8.356  & 5.701 \\
        BS    & 80.360 & 65.047 & 61.098 & 38.773 & 25.817 & 3.906 \\
        SS    & 82.069 & 68.203 & 62.073 & 45.178 & 31.936 & 4.47 \\
        \bottomrule
        \end{tabular}%
    }
    \captionof{table}{Per-class IoU (\%) for the STPLS3D dataset using RandLA-Net.}
    \label{tab:STPLS3D_per_class_results_RandLA}
\end{minipage}

\begin{table*}[t]
\begin{center}
    \resizebox{1.0\linewidth}{!}{%
        \begin{tabular}{l|ccccccccccccc}
        \toprule
        Method  & \textit{ceiling} & \textit{floor} & \textit{wall} & \textit{beam} & \textit{column} & \textit{window} & \textit{door} & \textit{chair} & \textit{table} & \textit{bookcase} & \textit{sofa} & \textit{board} & \textit{clutter} \\
        \midrule
        uni   & 93.636 & 98.515 & 80.843 & 0.000 & 22.259 & 44.236 & 61.458 & 87.107 & 79.278 & 70.984 & 64.165 & 60.649 & 57.433 \\
        invf  & 92.212 & 98.305 & 80.333 & 0.000 & 25.148 & 46.485 & 61.312 & 87.132 & 79.381 & 71.166 & 70.813 & 60.570 & 55.048 \\
        cb    & 92.728 & 98.405 & 80.571 & 0.000 & 24.341 & 44.645 & 62.272 & 87.440 & 78.390 & 71.073 & 67.183 & 62.709 & 56.127 \\
        invl  & 93.443 & 98.486 & 80.786 & 0.000 & 24.058 & 45.156 & 59.885 & 87.778 & 79.509 & 71.855 & 67.775 & 60.852 & 58.569 \\
        invp  & 93.010 & 98.415 & 80.546 & 0.000 & 23.339 & 46.768 & 60.452 & 87.024 & 78.933 & 70.266 & 60.969 & 61.074 & 57.592 \\
        comf  & 92.944 & 98.410 & 81.089 & 0.000 & 21.425 & 46.484 & 66.631 & 87.458 & 79.543 & 71.006 & 63.212 & 61.433 & 56.755 \\
        \hline
        FL    & 92.248 & 98.405 & 79.705 & 0.000 & 22.489 & 45.381 & 60.079 & 87.715 & 79.163 & 69.916 & 67.915 & 61.801 & 56.430 \\
        LDAM  & 92.706 & 98.473 & 80.670 & 0.000 & 24.218 & 45.179 & 61.765 & 87.993 & 78.938 & 71.449 & 63.680 & 61.173 & 56.468 \\
        LADJ  & 92.779 & 98.392 & 81.497 & 0.000 & 24.403 & 49.590 & 62.383 & 87.270 & 78.890 & 71.815 & 64.734 & 63.405 & 56.239 \\  
        BS    & 93.165 & 98.387 & 82.492 & 0.000 & 28.040 & 54.397 & 63.871 & 86.873 & 78.086 & 71.117 & 66.885 & 62.996 & 55.472 \\
        SS    & 92.947 & 98.459 & 80.608 & 0.000 & 21.814 & 48.496 & 62.857 & 87.806 & 78.637 & 70.714 & 67.919 & 62.051 & 55.672 \\
        \bottomrule
        \end{tabular}%
    }
\end{center}
\caption{Per-class IoU (\%) for the S3DIS dataset using KPConv.}
\label{tab:S3DIS_per_class_results_KPConv}
\end{table*}

\begin{table*}[b]
\begin{center}
    \resizebox{1.0\linewidth}{!}{%
        \begin{tabular}{l|ccccccccccccc}
        \toprule
        Method  & \textit{ceiling} & \textit{floor} & \textit{wall} & \textit{beam} & \textit{column} & \textit{window} & \textit{door} & \textit{chair} & \textit{table} & \textit{bookcase} & \textit{sofa} & \textit{board} & \textit{clutter} \\
        \midrule
        uni   & 93.116 & 97.028 & 80.367 & 0.000 & 17.390 & 57.660 & 36.838 & 78.216 & 84.665 & 55.844 & 70.799 & 71.421 & 54.524 \\
        invf  & 91.363 & 97.390 & 78.549 & 0.000 & 15.893 & 60.357 & 30.560 & 76.231 & 85.703 & 61.124 & 71.363 & 64.896 & 51.420 \\
        cb    & 91.821 & 97.074 & 79.904 & 0.000 & 26.317 & 61.150 & 33.409 & 78.371 & 81.896 & 75.646 & 70.521 & 64.509 & 50.663 \\
        invl  & 92.351 & 97.355 & 80.518 & 0.000 & 16.287 & 59.860 & 39.551 & 77.654 & 86.433 & 60.295 & 70.575 & 68.157 & 52.555 \\
        invp  & 92.722 & 97.734 & 80.037 & 0.000 & 22.035 & 59.819 & 41.352 & 78.766 & 86.838 & 72.210 & 71.187 & 72.921 & 52.304 \\
        comf  & 92.319 & 97.902 & 80.974 & 0.000 & 24.176 & 59.874 & 33.803 & 78.060 & 87.450 & 61.501 & 71.115 & 73.440 & 53.712 \\
        \hline
        FL    & 91.577 & 97.346 & 81.361 & 0.000 & 21.368 & 58.047 & 52.706 & 76.267 & 86.805 & 55.145 & 71.403 & 68.610 & 52.198 \\
        LDAM  & 92.158 & 96.902 & 81.553 & 0.000 & 28.788 & 59.396 & 50.455 & 77.356 & 88.658 & 66.383 & 72.024 & 73.208 & 54.194 \\
        LADJ  & 92.080 & 96.661 & 82.223 & 0.000 & 32.908 & 62.755 & 47.018 & 75.976 & 88.004 & 69.933 & 72.854 & 67.595 & 52.687 \\
        BS    & 91.241 & 97.327 & 81.789 & 0.000 & 23.710 & 61.296 & 43.184 & 78.355 & 87.483 & 67.940 & 71.363 & 63.855 & 52.973 \\
        SS    & 91.811 & 97.823 & 80.882 & 0.000 & 20.586 & 59.820 & 45.444 & 78.221 & 86.209 & 56.853 & 71.604 & 69.960 & 53.166 \\
        \bottomrule
        \end{tabular}%
    }
\end{center}
\caption{Per-class IoU (\%) for the S3DIS dataset using RandLA-Net.}
\label{tab:S3DIS_per_class_results_RandLA}
\end{table*}

\end{document}